# Cancer Vaccine Adjuvant Name Recognition from Biomedical Literature using Large Language Models

Hasin Rehana[1,2], Jie Zheng[3], Leo Yeh[3], Benu Bansal[2,4], Nur Bengisu Çam[5], Christianah Jemiyo[1], Brett McGregor[1], Arzucan Özgür[5], Yongqun He[3], and Junguk Hur[1,§]

[1]Department of Biomedical Sciences, University of North Dakota School of Medicine and Health Sciences, Grand Forks, North Dakota, 58202, USA

[2]School of Electrical Engineering & Computer Science, University of North Dakota, Grand Forks, North Dakota, 58202, USA

[3]Unit for Laboratory Animal Medicine, Department of Microbiology and Immunology, University of Michigan, Ann Arbor, Michigan, 48109, USA

[4]Department of Biomedical Engineering, University of North Dakota, Grand Forks, North Dakota, 58202, USA

[5]Department of Computer Engineering, Bogazici University, 34342 Istanbul, Turkey

§ Corresponding author: Junguk Hur, Department of Biomedical Sciences, University of North Dakota School of Medicine, 1301 N Columbia Rd. Grand Forks, ND 58202, Email: junguk.hur@med.und.edu

**Abstract**
**Motivation:** An adjuvant is a chemical incorporated into vaccines that enhances their efficacy by improving the immune response. Identifying adjuvant names from cancer vaccine studies is essential for furthering research and enhancing immunotherapies. However, the manual curation from the constantly expanding biomedical literature poses significant challenges. This study explores the automated recognition of vaccine adjuvant names using state-of-the-art Large Language Models (LLMs), specifically Generative Pretrained Transformers (GPT) and Large Language Model Meta AI (Llama).
**Methods:** We utilized two datasets: 97 clinical trial records from AdjuvareDB and 290 PubMed abstracts annotated with the Vaccine Adjuvant Compendium (VAC). Two LLMs, GPT-4o and Llama 3.2 were employed in zero-shot and few-shot learning paradigms with up to four examples per prompt. Prompts explicitly targeted adjuvant names, testing the impact of contextual information such as substances or interventions. Outputs underwent automated and manual validation for accuracy and consistency.
**Results:** GPT-4o consistently attained 100% Precision across all situations, while also exhibiting notable enhancements in Recall and F1-scores, particularly with the incorporation of interventions. On the VAC dataset, GPT-4o achieved a maximum F1-score of 77.32% with interventions, surpassing Llama-3.2-3B by approximately 2%. On the AdjuvareDB dataset, GPT-4o reached an F1-score of 81.67% for three-shot prompting with interventions, surpassing Llama-3.2-3B's maximum F1-score of 65.62%. These results highlight the critical role of contextual information in enhancing model performance, with GPT-4o demonstrating a superior ability to leverage this enrichment.
**Conclusion:** Our findings demonstrate that LLMs excel at accurately identifying adjuvant names, including rare and novel variations of naming representation. This study emphasizes the capability of LLMs to enhance cancer vaccine development by efficiently extracting insights from clinical trial data. Future work aims to broaden the framework to encompass a wider array of biomedical literature and enhance model generalizability across various vaccines and adjuvants.

**Availability:** Source code is available at https://github.com/hurlab/Vaccine-Adjuvant-LLM.

## 1. Introduction

Vaccine represents one of the most significant advancements in medical history, providing effective prevention and control of infectious diseases. Cancer immunotherapy has leveraged vaccine technologies to stimulate the immune system to recognize and destroy tumor cells. Vaccines have markedly diminished the effects of infectious diseases, with a historical trajectory commencing 500 years ago and culminating in the eradication of smallpox in 1980. The swift advancement of COVID-19 vaccines has emphasized the significance of vaccine innovation, showcasing improvements in laboratory methodologies that persist in saving millions of lives and providing critical insights for future pandemic preparedness (Saleh et al., 2021).

Extending these successes to non-viral tumors and improving overall cancer prevention strategies are necessary next steps in cancer vaccine development, which has made significant achievements in preventing virus-related cancers, especially with successful vaccines. However, there are still obstacles to overcome, such as identifying target antigens in at-risk individuals and inducing long-lasting immune responses (Ruzzi et al., 2025). These vaccines frequently incorporate adjuvants—substances that enhance the immunogenicity of antigens—to improve their efficacy. Identifying and characterizing potent adjuvants remains critical for advancing cancer vaccine development, as these components directly influence the magnitude and quality of immune responses.

Cancer vaccine development has undergone significant evolution over the past decades, with advancements in molecular biology, immunology, and bioinformatics contributing to novel approaches in vaccine design. A crucial aspect of cancer vaccines is the inclusion of adjuvants, which are pivotal in enhancing the immune response against weakly immunogenic tumor antigens. Traditional adjuvants, such as aluminum salts and oil-in-water emulsions, have been widely studied and utilized in prophylactic vaccines. However, the unique immunosuppressive microenvironment of tumors necessitates the development of more potent and specialized adjuvants tailored for therapeutic cancer vaccines. Finding the best combinations of well-known or currently available adjuvants could aid in developing new ones for cancer immunotherapy. This superiority of immunostimulatory or immunomodulatory adjuvant combinations over individual use has been well-demonstrated (Temizoz et al., 2016).

Understanding vaccine adjuvants and their processes to boost T cell responses and improve clinical outcomes for cancer patients is crucial, especially as cancer vaccines are considered potential partners with immunotherapies such as T cell checkpoint inhibition (Khong and Overwijk, 2016). Artificial Intelligence (AI) is transforming the expensive and time-consuming drug development process by improving patient selection and streamlining target discovery, especially in oncology. Although obstacles exist, such as limited data accessibility and a lack of trained staff, AI can enhance cancer vaccination effectiveness through novel adjuvant design and by improving customized therapy (Zhang et al., 2024).

In recent years, computational approaches have emerged as powerful tools for advancing cancer vaccine research. Machine learning (ML) techniques have been applied to predict immunogenic epitopes, analyze clinical trial outcomes, and identify novel antigens. Kumar et al. highlight how advancements in AI, particularly ML and computational modeling, have enabled the precise prediction and optimization of neoantigens, improved vaccine design, and facilitated the creation of personalized cancer vaccines (Kumar et al., 2024).

Studies have demonstrated the efficacy of models like Generative Pretrained Transformers (GPT) and Large Language Model Meta AI (Llama) in tasks such as named entity recognition and relation extraction, which are critical for understanding the interplay between vaccine components and their immunological outcomes (Palepu et al., 2024). A novel annotation schema for oncology information was created and evaluated using large language Models (LLMs), demonstrating that although GPT-4 exhibited superior performance in extracting comprehensive oncological histories from clinical notes, substantial enhancements are still required for dependable use in clinical research and patient care documentation (Sushil et al., 2024).

For instance, Ferber et al. illustrates that localized fine-tuning of Llama models via the QLoRA algorithm can proficiently produce physician letters in radiation oncology, achieving significant therapeutic advantages and efficiency with minimal computational resources (Ferber et al., 2024). Using LLMs, authors developed an automated pipeline that accurately matches cancer patients to clinical trials, identifying 93.3% of relevant trials and achieving matches in 92.7% of cases. This improves the process of matching patients to trials and may be better than qualified medical professionals (Hou et al., 2024).

Although research leveraging LLMs for cancer vaccine adjuvant recognition is still nascent, a few pioneering studies underscore its potential. For instance, VaxLLM fine-tuned a large LLM to annotate vaccine components, including adjuvants, in Brucella vaccines (Li et al., 2024). Similar methodologies could be adapted to cancer vaccine research, focusing on extracting adjuvant-specific entities from clinical trial data. Studies on oncology guidelines and personalized oncology have also explored LLMs for tasks like zero-shot learning, achieving notable accuracy improvements through few-shot training (Benary et al., 2023).

While these applications are not directly centered on cancer vaccine adjuvants, they highlight the broader utility of LLMs in biomedical research and their promise to advance this niche area. This manuscript investigates the application of LLMs for recognizing cancer vaccine adjuvant names from clinical trial data. By harnessing the advanced NLP capabilities of LLMs, this study proposes a systematic framework for extracting adjuvants referenced in cancer vaccine trials. This approach not only facilitates a more comprehensive understanding of the role of adjuvants in cancer immunotherapy but also underscores the potential of LLMs to advance biomedical research through data-driven insights.

## 2. Adjuvant Databases and Resources

The discovery and development of vaccine adjuvants rely heavily on curated databases and resources that compile critical information on adjuvant properties, usage, and safety. Several prominent databases have emerged as invaluable tools to support researchers in accessing and analyzing adjuvant-related data:

**AdjuvareDB:** AdjuvareDB is a comprehensive web-based database that compiles information on candidate adjuvants in clinical use (Ren et al., 2024). It provides detailed records of adjuvant composition, function, and other attributes, serving as a valuable resource for understanding their applications in immunotherapy.

**Vaxjo:** Vaxjo is a centralized web-based database and analysis platform designed to curate, store, and analyze vaccine adjuvants and their roles in vaccine development (Sayers et al., 2012). The database includes detailed information such as adjuvants names, components, structure, appearance, storage conditions, preparation methods, function, safety, and associations with specific vaccines. This robust database facilitates the exploration of adjuvant characteristics and their applications across diverse vaccine platforms.

**Vaccine Adjuvant Compendium (VAC):** The VAC database (https://vac.niaid.nih.gov/) focuses on providing comprehensive records for each adjuvant, including points of contact for intellectual property (IP) holders. Each entry in VAC encompasses information such as the Vaccine Ontology ID, detailed properties of the adjuvant, preclinical and clinical usage data, associated publications, product grade, and available formulations. By offering these comprehensive insights, VAC aids researchers in identifying and leveraging adjuvants for vaccine development.

By centralizing and standardizing critical information on adjuvants, these databases are advancing vaccine research. They enable researchers to make informed decisions when selecting adjuvants for experimental and clinical applications, thus accelerating the development of next-generation vaccines.

## 3. Large Language Model (LLM)

LLMs are machine learning models designed to process and generate human-like text by learning patterns and relationships within vast textual datasets (Thirunavukarasu et al. 2023). These models, such as Generative Pretrained Transformers (GPT) and Llama, utilize transformer-based architectures, enabling them to handle complex linguistic tasks with high accuracy. Their capabilities extend across a broad spectrum of natural language processing (NLP) applications, including named entity recognition (NER), text summarization, translation, and contextual understanding.

In the biomedical domain, LLMs have demonstrated immense potential in addressing challenges posed by unstructured data. Tasks such as extracting meaningful information from scientific literature, annotating clinical data, and identifying relationships between entities have greatly benefited from LLM-driven automation. Their ability to contextualize and synthesize information makes them particularly valuable in fields requiring the analysis of large and complex datasets, such as cancer vaccine development.

One of the notable applications of LLMs is in named entity recognition, where these models are fine-tuned to identify and categorize specific entities, such as genes, proteins, or adjuvants, from text. For instance, studies leveraging fine-tuned LLMs, like VaxLLM, have shown the effectiveness of such approaches in annotating vaccine-related components, paving the way for similar applications in cancer vaccine research. Moreover, few-shot and zero-shot learning capabilities further enhance the utility of LLMs, enabling them to generalize to new tasks with minimal labeled data (Li et al., 2024).

The evaluation of models such as GPT and BERT for protein-protein interaction identification (Rehana et al., 2024b) and nested named entity recognition using multilayer BERT-based architectures (Rehana et al., 2024a) further showcases the adaptability of LLMs for nuanced and complex biomedical tasks. These advancements underline the transformative potential of LLMs in automating and scaling research efforts, especially in specialized domains like cancer vaccine adjuvant discovery.

## 4. Methods

We have employed two LLMs, namely Llama and GPT, in this research. **Figure 1** illustrates the overall structure of our methodology.

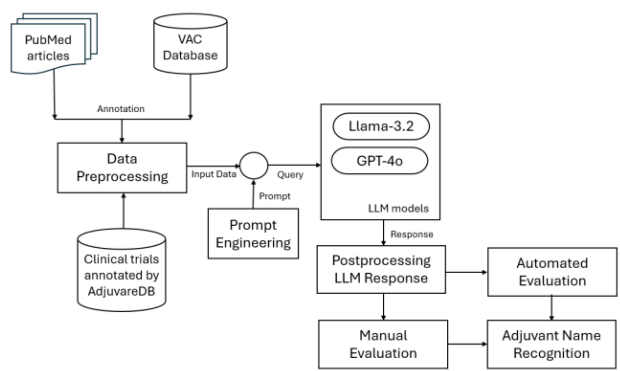

**Figure 1. Methodology for Adjuvant Name Recognition.** Overview of the methodology with processing pipelines.

### 4.1 Llama

The Llama is a family of open-source large-scale language models developed by Meta AI, designed to achieve cutting-edge performance across various benchmarks (Touvron et al., 2023). The initiative prioritizes openness and efficiency, allowing the academic community to use high-performing models without relying on proprietary datasets. Key contributions include efficient scaling, which challenges the concept that only the largest models get the best results.

Llama takes advantage of publicly available datasets such as CommonCrawl, Wikipedia, and GitHub repositories, which makes the models powerful and compatible with open-source platforms. Llama models are available in various configurations, including lightweight versions. The lightweight models (e.g. Llama 3.2 1B, Llama 3.2 3B) are optimized for devices with constrained resources without significantly compromising accuracy. However, the 1B Llama model struggles to follow the instruction to provide consistently structured output format in our study (data not shown). Llama models demonstrate competitive performance across tasks, including common sense reasoning, question answering, and reading comprehension. They also have impressive zero-shot and few-shot learning skills. Furthermore, Llama models are trained to utilize energy-efficient techniques by prioritizing memory optimization and minimizing activation recomputation. This technique is consistent with the ideas of sustainable AI, lowering computational resource demands while maintaining good performance.

Our research utilized the Llama 3.2 3B model. The temperature parameter was set to 0.0001 to balance creativity and coherence in responses. The maximum token limit was set to 100 to ensure concise output. The model was configured to perform in both zero-shot and few-shot learning setups, using zero to four-shot examples for task adaptation. The transformer-based design of Llama, with multi-head self-attention mechanisms, enables adequate contextual understanding across input text, regardless of positional relationships.

### 4.2 GPT

GPT models represent a groundbreaking class of LLMs designed to process and generate natural language text (Radford, 2018). Developed by OpenAI, GPT models, such as GPT-3 and GPT-4, have set a new standard for natural language processing (NLP) tasks, owing to their remarkable ability to understand context, generate coherent responses, and perform complex linguistic reasoning. Early studies demonstrate GPT models' potential in successfully addressing complex biomedical challenges, such as drug discovery optimization and protein-protein interaction identification, with high accuracy and efficiency. Their transformer-based architecture allows them to process large volumes of unstructured data, making them invaluable for applications like literature mining and clinical data analysis.

Our study employed GPT-4o (Hurst et al., 2024) for cancer vaccine adjuvant identification with a temperature value of 0.0001 to generate precise and controlled outputs. The model was utilized in a zero-shot and few-shot learning paradigm, incorporating zero to four-shot examples when necessary to enhance task-specific performance. One key strength of GPT models is their ability to perform few-shot and zero-shot learning, a technique in which the model can learn new tasks with minimal or no prior exposure to labeled data.

Despite challenges such as the potential risk of generating hallucinated information and the necessity for rigorous output validation, GPT models' scalability, versatility, and seamless integration with computational tools position them as crucial assets in accelerating advancements in cancer vaccine development, drug discovery, and other critical areas of biomedical research.

### 4.3 Data Preprocessing

This study utilized datasets for cancer vaccine adjuvant name recognition from two primary sources: gold standard annotated clinical trial records from the AdjuvareDB website (http://tmliang.cn/adjuvaredb/) and PubMed abstracts annotated in the Vaccine Adjuvant Compendium (VAC) database (https://vac.niaid.nih.gov/). The AdjuvareDB dataset included **97** trials manually annotated by the AdjuvareDB team. The VAC dataset comprised **290** abstracts from clinical and preclinical studies collected from PubMed.

The yearly distribution of the 290 abstracts from the VAC dataset illustrated in **Figure 2** illustrates a significant increase in cancer vaccine adjuvant-related studies in recent years. This trend underscores the growing complexity and scale of curating cancer vaccine-related information manually.

To meet the urgent need for scalable solutions in cancer vaccine research, this study automated the identification of adjuvant names using advanced LLMs. A detailed breakdown of the datasets, including annotation sources, is presented in **Table 1**. These two datasets have formed a robust foundation for our research to develop and evaluate automated adjuvant name recognition methods.

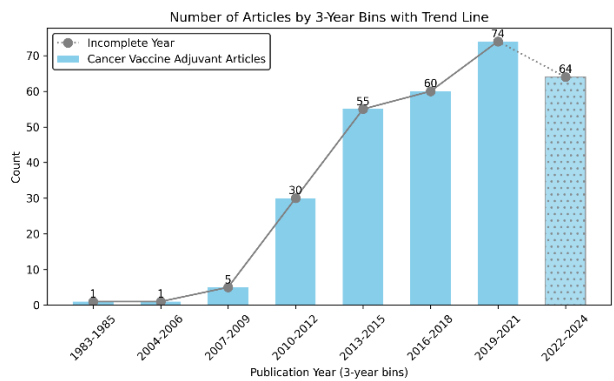

**Figure 2. Year-wise distribution of the PubMed articles in the VAC dataset.**

**Table 1. Datasets.** Overview of datasets used for cancer vaccine adjuvant name recognition, detailing sources, annotations, specifications, and target variables.

| Dataset type & Source | Annotation Source | Number of Entries | Dataset Specifications |
|---|---|---|---|
| Clinical trial dataset (clinicalTrials.gov) | AdjuvareDB website | 97 | Title: The official title of each clinical trial. Brief Summary: A concise description of the trial objectives and design. Interventions: Detailed descriptions of the trial interventions. |
| PubMed abstract dataset | VAC | 290 | Title: The title of the manuscript. Abstract: Abstract of the Manuscript. Substances: A detailed list of substances mentioned in the PubMed website |

Our study aimed to create a comprehensive, high-quality resource for cancer vaccine adjuvant research, supporting robust model development and evaluation. Both clinical trial records and research articles were used for this research, providing a comprehensive and high-quality resource for cancer vaccine adjuvant research.

### 4.4 Prompt Engineering

We designed a series of carefully crafted prompts tailored for each input type to extract target cancer vaccine adjuvant names effectively from PubMed abstracts and clinical trial datasets. The prompts explicitly defined the goal as identifying and extracting specific vaccine adjuvant names mentioned in the input text.

The significant parts of the prompt are task specification, key instructions, output format, and task input. Each query was designed to process one article or trial dataset at a time to avoid data mixing. Input data contained the unique identifier (PMID/NCT Number), title, article/trial data, and, optionally, substances/interventions.

To evaluate the impact of additional context, for PubMed abstracts, prompts were tested with and without the inclusion of substances to determine their role in enhancing extraction accuracy. Similarly, for clinical trials, prompts were evaluated with and without the list of interventions to assess their influence on identifying relevant adjuvant names. The zero-shot prompts for both datasets are detailed in **Figures 3 and 4**, which illustrate the basic structure and configuration for each setting.

```
Task: Extract specific vaccine adjuvant names from the provided article data. Each input consists of a PMID (unique identifier), article title, abstract, and substances separated by tabs. Your task is to identify explicit mentions of adjuvants and pair each with the corresponding PMID.
---
### Key Instructions:
Definition of Adjuvants:
Adjuvants are substances that enhance the body's immune response to an antigen. Focus on identifying components explicitly described as adjuvants or known to act as adjuvants in the article.
Avoid Generic Terms:
Ignore generic mentions of "adjuvant" unless accompanied by a specific name or descriptor (e.g., "Hepatitis B Core Antigen (HBcAg)" described as a Th1 adjuvant).
Exact Names:
Use the exact wording for adjuvants as mentioned in the article. Avoid paraphrasing or adding external adjuvant names not found in the text.
Maximum Outputs:
Limit to two distinct adjuvant names per article. Avoid duplicate rows.
---
### Output Format:
Produce a TSV (tab-separated values) Figure with the following columns:
PMID    Adjuvant Name
If multiple adjuvants are identified in a single article, each adjuvant should be listed on a separate row under the same PMID. Provide the adjuvant name exactly as mentioned in the article. Limit the output to a maximum of three distinct adjuvant names per article. Avoid duplicate rows. At the end of the output, include a line with the word "Done" to indicate the completion of processing.
---
### Task Input:
…………..
PMID           Article
PMID_NNN    Title: TTT. Abstract: AAA. Substances: SSS.
```

**Figure 3. Prompt for PubMed Abstract Dataset.**

```
Task: Extract specific vaccine adjuvant names from the provided clinical trial data. Each input consists of an NCT Number (unique identifier), trial title, brief description, and interventions separated by tabs. Your task is to identify explicit mentions of adjuvants and pair each with the corresponding NCT Number.
---
### Key Instructions:
Definition of Adjuvants:
Adjuvants are substances that enhance the body's immune response to an antigen. Focus on identifying components explicitly described as adjuvants or known to act as adjuvants in the article.
Avoid Generic Terms:
Ignore generic mentions of "adjuvant" unless accompanied by a specific name or descriptor (e.g., "Hepatitis B Core Antigen (HBcAg)" described as a Th1 adjuvant).
Exact Names:
Use the exact wording for adjuvants as mentioned in the trial data. Avoid paraphrasing or adding external adjuvant names not found in the text.
Maximum Outputs:
Limit to two distinct adjuvant names per trial. Avoid duplicate rows.
---
### Output Format:
Produce a TSV (tab-separated values) table with the following columns:
NCT Number  Adjuvant Name
If multiple adjuvants are identified in a single trial, each adjuvant should be listed on a separate row under the same NCT Number. Provide the adjuvant name exactly as mentioned in the trial. Limit the output to a maximum of three distinct adjuvant names per trial. Avoid duplicate rows. At the end of the output, include a line with the word "Done" to indicate the completion of processing.
---
### Task Input:
…………..
NCT Number    Trial Data
NCT_NNN      Title: TTT. Brief Description: DDD. Interventions: III.
```

**Figure 4. Prompt for Clinical Trial Dataset.**

### 4.5 Postprocessing

We have employed a systematic postprocessing approach to extract the meaningful response from the LLM models. The process involved data cleaning, removing duplicates, formatting standardization and ensuring the response was complete. We have removed any details other than the tab-delimited table and duplicate rows to avoid unnecessary redundancy. We have ensured outputs adhered to the specified tab-separated values (TSV) format, including the unique identifier (PMID for PubMed and NCT ID for clinical trials) and corresponding adjuvant names. We have checked for a "Done" marker in the output to signal the end of processing for each input. This organized step after response extraction ensured that the data met high standards of accuracy and consistency. This made it possible to use the data reliably in later analyses and helped reach the goal of automatically finding cancer vaccine adjuvants in biomedical literature.

### 4.6 Performance Evaluation

The study evaluated a model's performance using Precision, Recall, and F1 scores. Precision measures the accuracy of true positive outputs, by assessing how well the model minimized nonspecific or spurious results. A higher Precision score indicates that the model minimized nonspecific outputs and focuses on relevant results. Recall measures the model's ability to identify all relevant instances, including missed potential positives. A higher Recall score indicates that the model captures a larger proportion of true positive instances, even with nonspecific outputs. The F1 score, the harmonic mean of Precision and Recall, offers a balanced performance measure. A higher F1 score indicates a model that effectively balances Precision and Recall, ensuring output accuracy and comprehensiveness. The equations for these metrics in this research are,

$$Precision = \frac{True\ Positive - Nonspecific Output}{True\ Positive} \quad \ldots(1)$$

$$Recall = \frac{True\ Positive}{Total\ Identification + Missed\ Instances} \quad \ldots(2)$$

$$F1\ Score = \frac{2 \times Precision \times Recall}{Precision + Recall} \quad \ldots(3)$$

These metrics collectively provide a robust framework for evaluating the model's accuracy, comprehensiveness, and balance in its outputs. By analyzing these metrics, the study identified areas for targeted improvement, ensuring alignment with the intended objectives.

The evaluation process employed a combination of automated and manual validation methods to ensure the accuracy and reliability of the evaluation scores. Automated validation served as the initial step in the pipeline, applying an exact-match (case insensitive) criterion to compare the model's outputs against a curated dictionary of predefined mappings. The dictionary, meticulously compiled and validated in advance, acted as the reference for determining correctness. Automated validation was efficient in quickly identifying outputs that matched the expected results. However, its limitations in handling ambiguous, context-dependent, or nuanced cases, necessitated further scrutiny through manual validation.

Mismatched cases identified during the automated process were subjected to manual validation to address the limitations. A team of six domain experts thoroughly reviewed each mismatched output to determine its correctness. At least two validators reviewed each case independently, ensuring that each instance was examined from multiple perspectives, reducing the likelihood of oversight or bias. In cases where the two initial validators disagree, the instance was forwarded to a third validator. The third validator reviewed the case independently and provided the final decision, resolving discrepancies and ensuring a fair and accurate validation. Findings are carefully documented throughout the manual validation process, including the reasons for disagreements and their resolution.

### 5. Results and Discussion

A few examples of GPT-4o and LlaMA-3.2 outputs are listed in **Table 2**.

**Table 2. LLM outputs on cancer vaccine name identification.**

| Input Type | Model | Output |
|---|---|---|
| Abstract | Llama-3.2 | PMID  Adjuvant Name<br>PMID_26407920 Advax<br>PMID_26407920 Delta inulin<br>Done |
| Clinical Trial | Llama-3.2 | NCT Number Adjuvant Name<br>NCT00471471 GM-CSF<br>NCT00471471 Incomplete Freund's adjuvant<br>NCT00471471 CpG 7909 |
| Clinical Trial | GPT-4o | ### Output:<br>```<br>PMID  Adjuvant Name<br>PMID_25367751 GLA-SE<br>PMID_25367751 Squalene oil-in-water emulsion (SE)<br>Done<br>``` |
| Abstract | GPT-4o | ### Output:<br>```<br>NCT Number Adjuvant Name<br>NCT00694551 Poly IC-LC<br>NCT00694551 Hiltonol<br>Done<br>``` |

**Table 3** provides insights into the performance of GPT-4o and Llama-3.2-3B models across automated and manual validation processes on the VAC dataset. The GPT-4o model achieved consistent 100% Precision across all shots except for two-shot prompting without setting substances. Recall began at 34.05% for zero-shot scenarios, with an F1-score of 50.80%. It steadily improved to a Recall of 46.69% and an F1-score of 63.66% at four shots.

We manually validated 928 initial mismatches that consistently appeared in at least two independent runs during our experiments. Each case was reviewed by two independent experts, resulting in 144 cases of disagreement. These disagreements were subsequently reviewed by a third validator, who provided the final verdict.

With manual validation, Recall improved further from 48.28% to 60.94%, and F1-score increased from 65.08% to 75.73%. These results highlight the limitations of the ability of LLM models to properly format and deliver the output as instructed.

With the addition of substances to the prompts along with texts, Recall slightly improved compared to the Without Substances case. Recall rose from 33.07% in zero-shot scenarios to 47.05% at four shots, while the F1 score followed a similar trend, increasing from 49.69% to 63.99%. Manual validation further enhanced Recall, increasing from 50.84% to 63.03%, and the F1 score improved from 67.40% to 77.32%. These findings suggest that the inclusion of substances resulted provided contextual enrichment, which benefited GPT-4o's performance.

The Llama-3.2-3B model exhibited slightly lower Precision (less than ~1%) than GPT-4o for some of the few shot prompts. However, Recall and F1 scores were significantly lower, starting at 24.21% and 38.58%, respectively, in the zero-shot setting. Even with four shots, the Recall only improved to 34.58%, and the F1 score reached 51.39%.

Manual validation revealed notable improvement over automated validation for Llama-3.2-3B model. Recall increased from 39.09% to 49.82%, with corresponding F1 scores improving from 56.05% to 66.50%. However, Llama-3.2-3B still lagged behind GPT-4o in terms of both metrics. Introducing substances also led to a noticeable improvement in Recall and F1 scores. Recall rose from 25.05% to 35.35%, and the F1 score improved from 40.00% to 55.44%. However, these values remained lower than those achieved by GPT-4o in the same conditions. Manual validation further boosted the model's performance, increasing Recall from 46.12% in the zero-shot scenario to 60.32% at four shots, with the F1 score improving from 62.94% to 75.25%. While the performance gap narrowed in some instances, GPT-4o generally outperformed Llama-3.2-3B for the VAC dataset.

**Table 3. Comparative Results on VAC dataset.** Direct comparison of GPT-4o, Llama-3.2-1B, and Llama-3.2-3B models, highlighting the differences in performance metrics (Precision, Recall, and F1-score) with and without interventions.

| | | Automated Validation | | | Manual Validation | | |
|---|---|---|---|---|---|---|---|
| **Models** | Shots | P (%) | R (%) | F1 (%) | P (%) | R (%) | F1 (%) |

| Models | Shots | P (%) | R (%) | F1 (%) | P (%) | R (%) | F1 (%) |
|---|---|---|---|---|---|---|---|
| GPT-4o Without Substances | 0 | **100.00** | 34.05 | 50.80 | **100.00** | 48.28 | 65.08 |
|  | 1 | **100.00** | 38.30 | 55.39 | **100.00** | 54.40 | 70.47 |
|  | 2 | 99.80 | 40.03 | 57.14 | 99.86 | 56.14 | 71.87 |
|  | 3 | **100.00** | 45.48 | 62.52 | **100.00** | 60.44 | 75.34 |
|  | 4 | **100.00** | 46.69 | 63.66 | **100.00** | 60.94 | 75.73 |
| GPT-4o With Substances | 0 | **100.00** | 33.07 | 49.69 | **100.00** | 50.84 | 67.40 |
|  | 1 | **100.00** | 38.62 | 55.72 | **100.00** | 56.53 | 72.22 |
|  | 2 | **100.00** | 39.73 | 56.87 | **100.00** | 57.26 | 72.82 |
|  | 3 | **100.00** | 46.27 | 63.27 | **100.00** | 61.75 | 76.35 |
|  | 4 | **100.00** | **47.05** | **63.99** | **100.00** | **63.03** | **77.32** |
| Llama-3.2-3B-Instruct Without Substances | 0 | 98.00 | 24.21 | 38.58 | 99.01 | 39.09 | 56.05 |
|  | 1 | **100.00** | 32.81 | 49.41 | **100.00** | 47.80 | 64.68 |
|  | 2 | **100.00** | 33.37 | 50.04 | **100.00** | 50.00 | 66.67 |
|  | 3 | 99.52 | 35.75 | 52.60 | 99.67 | 51.98 | 68.33 |
|  | 4 | **100.00** | 34.58 | 51.39 | **100.00** | 49.82 | 66.50 |
| Llama-3.2-3B-Instruct With Substances | 0 | 99.16 | 25.05 | 40.00 | 99.09 | 46.12 | 62.94 |
|  | 1 | **100.00** | 33.75 | 50.47 | **100.00** | 56.64 | 72.32 |
|  | 2 | **100.00** | 35.59 | 52.83 | **100.00** | 57.43 | 72.96 |
|  | 3 | 99.50 | 39.54 | 56.59 | 99.68 | 61.19 | 75.83 |
|  | 4 | **100.00** | 35.35 | 55.44 | **100.00** | **60.32** | **75.25** |

**Table 4. Comparative Results on AdjuvareDB dataset annotated by AdjuvareDB team.** Direct comparison of GPT-4o, Llama-3.2-1B, and Llama-3.2-3B models, highlighting the differences in performance metrics (Precision, Recall, and F1-score) with and without interventions. Includes statistical significance of observed improvements.

|  |  | Automated Validation |  |  |
|---|---|---|---|---|
| **Models** | **Shots** | P (%) | R (%) | F1 (%) |
| GPT-4o (Without Interventions) | 0 | 97.73 | 50.58 | 66.65 |
|  | 1 | **100.00** | 60.75 | 75.58 |
|  | 2 | **100.00** | 63.90 | 77.98 |
|  | 3 | **100.00** | 62.63 | 77.02 |
|  | 4 | **100.00** | 63.48 | 77.66 |
| GPT-4o (With Interventions) | 0 | **100.00** | 62.31 | 76.78 |
|  | 1 | **100.00** | 64.62 | 78.51 |
|  | 2 | **100.00** | 67.50 | 80.59 |
|  | 3 | **100.00** | **69.02** | **81.67** |
|  | 4 | **100.00** | 66.67 | 80.00 |
| Llama-3.2-3B-Instruct (Without Interventions) | 0 | **100.00** | 25.97 | 41.11 |
|  | 1 | **100.00** | 27.09 | 42.63 |
|  | 2 | 99.19 | 29.57 | 45.55 |
|  | 3 | **100.00** | 37.46 | 54.50 |
|  | 4 | **100.00** | 34.63 | 51.44 |
|  | 0 | 97.66 | 37.32 | 54.00 |
| Llama-3.2-3B-Instruct (With Interventions) | 1 | 98.55 | 46.73 | 63.39 |
|  | 2 | 99.54 | 47.41 | 64.22 |
|  | 3 | **100.00** | 48.48 | 65.31 |
|  | 4 | **100.00** | **48.84** | **65.62** |

**Table 4** provides a comparative analysis of GPT-4o and Llama-3.2-3B models on the Clinical Trial dataset annotated by the AdjuvareDB team. This result also indicates that GPT-4o outperformed Llama-3.2-3B, particularly in Recall and F1-scores, while both models demonstrated consistently high Precision. GPT-4o achieved a maximum F1-score of 81.67% with three-shot interventions, markedly surpassing Llama-3.2-3B, which reached 65.62% under similar circumstances.

The inclusion of interventions had a clear impact, especially for Recall, where GPT-4o demonstrated a more significant ability to leverage this contextual enrichment, achieving a Recall of 69.02% compared to Llama-3.2-3B's 48.84%. These findings highlight GPT-4o's strength and efficacy in identifying adjuvant names from clinical trial data. With interventions, GPT-4o maintained 100% Precision across all shots, with Recall improving significantly as more examples were incorporated. While Llama-3.2-3B also showed consistently high Precision, its Recall values were lower across the board. Interventions contributed to better Recall and F1 scores for both models, but GPT-4o consistently outperformed Llama-3.2-3B in all metrics, particularly in Recall and F1-scores.

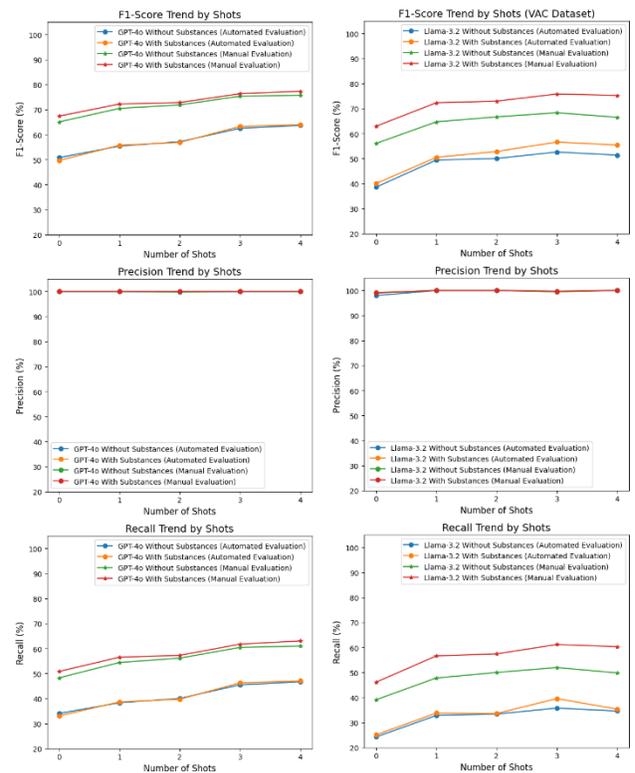

**Figure 5. Comparative Results on VAC dataset.**

**Figure 5** clearly highlights the improvement of F1 score and Recall as the number of few-shot examples and contextual information increased. Providing more examples enabled the model to better classify and identify substances, offering a stronger foundation for accurate prediction. Both models maintained near-perfect Precision across all settings, indicating strong reliability in identifying relevant cases. The optimal F1-score for GPT-4o was achieved at three-shots with interventions (81.67%), while for Llama-3.2-3B, it plateaued at 65.62% with four shots and interventions on VAC dataset. A similar trend was observed in Figure 6 as well for AdjuvareDB dataset.

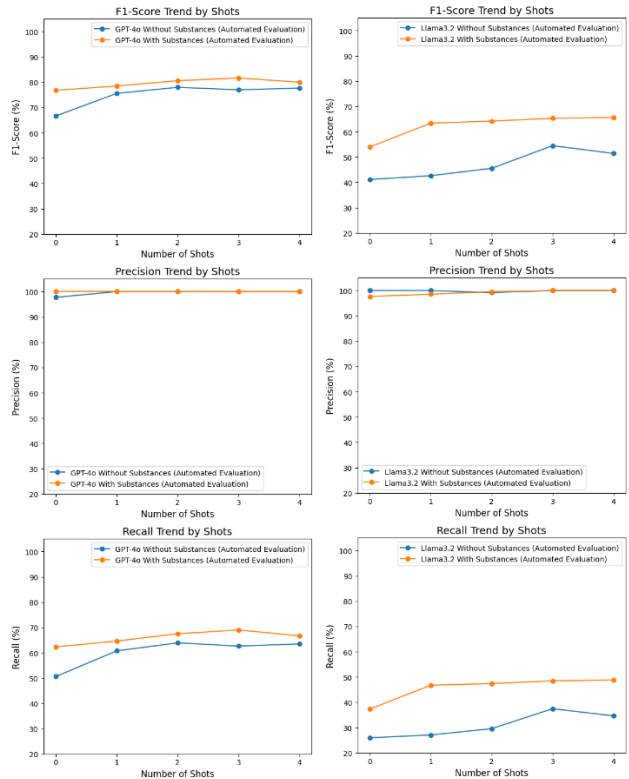

**Figure 6. Comparative Results on AdjuvareDB dataset.**

Automated validation provided speed and consistency in identifying exact matches, while manual validation introduced the expertise and judgment necessary for handling edge cases and ambiguities. This meticulous two-step validation process enhanced the reliability and comprehensiveness of the evaluation scores, ensuring they accurately reflect the model's performance and provided actionable insights for further refinement and improvement.

Besides, during the manual review, we found that some of the adjuvant names identified by the LLM models were actually correct or valid but not included in the "gold standard" dataset. In the current evaluation settings, such valid adjuvants were considered incorrect simply because they were not part of the gold standard reference. Our study suggests there is a gap in the dataset's comprehensiveness, and LLMs are good at identifying those valid but unlisted adjuvants.

The findings of this study underscore the potential of LLMs in addressing domain-specific challenges in cancer vaccine research. Our results demonstrate that models like GPT-4 and Llama are highly effective at accurately identifying cancer vaccine adjuvant names, even in large and diverse datasets. This systematic design of prompts provided a robust framework for extracting adjuvants, allowing us to evaluate the effect of contextual information and other prompt variations in prompt configurations. Clear, concise instructions and structured outputs ensured precise and consistent results across datasets.

The need for automation in this field is evident, as manual curation is increasingly unable to cope with the rapid growth of cancer vaccine research. The superior performance of fine-tuned LLMs, particularly in terms of Recall and F1 score, highlights their capability to distinguish adjuvant names from other biomedical terms, reduce false positives, and capture rare and novel entities. However, the results also revealed the importance of manual validation in addressing formatting and normalization issues, emphasizing that human oversight remains critical in achieving optimal accuracy.

While this study focused on cancer vaccine adjuvants, the framework developed here could support similar tasks in biomedical research that involve unstructured textual datasets. For instance, this approach may prove beneficial in streamlining the extraction of other vaccine components or clinical trial details. By addressing the challenges of scale and complexity in biomedical data, LLM-based frameworks offer a path toward more efficient and accurate information retrieval processes, particularly in highly specialized domains.

## 6. Future Direction

In the future, our research will extend beyond cancer vaccine adjuvants to include those targeting infectious diseases, addressing broader public health challenges. We plan to explore larger variants of Llama models (e.g. Llama-3.3 70B) as well as other open-access LLMs to enhance the flexibility and efficiency of our methodologies. Additionally, we aim to integrate vaccine ontology into our data preprocessing and fine-tuning pipelines, ensuring improved semantic accuracy and a deeper contextual understanding of adjuvants. By refining model generalizability and expanding datasets, we aim to create robust framework that contribute to advancing vaccine research and innovation on a global scale.

## Acknowledgments

The study was supported by the U.S. National Institute of Allergy and Infectious Disease (U24AI171008 to Y.H. and J.H.).